\documentclass[format=sigconf]{acmart}
\usepackage{amsmath}
\usepackage{microtype}
\usepackage{graphicx}
\usepackage{subfigure}
\usepackage{booktabs}
\usepackage{float}
\usepackage{hyperref}

\usepackage{algorithm, algorithmic}

\usepackage{tabularx}
\usepackage{makecell}
\usepackage{multirow}

\begin{document}

\title{Metrics and methods for a systematic comparison of fairness-aware machine learning algorithms}

\author{Gareth P. Jones}
\affiliation{%
\institution{Experian DataLabs UK\&I and EMEA}
\city{London}
\country{United Kingdom}
}

\author{James M. Hickey*}\thanks{*Contact author.}
\email{James.Hickey@experian.com}
\affiliation{%
\institution{Experian DataLabs UK\&I and EMEA}
\city{London}
\country{United Kingdom}
}

\author{Pietro G. Di Stefano}
\affiliation{%
\institution{Experian DataLabs UK\&I and EMEA}
\city{London}
\country{United Kingdom}
}

\author{Charanpal Dhanjal}
\affiliation{%
\institution{Experian DataLabs UK\&I and EMEA}
\city{London}
\country{United Kingdom}
}

\author{Laura C. Stoddart}
\affiliation{%
\institution{Experian DataLabs UK\&I and EMEA}
\city{London}
\country{United Kingdom}
}

\author{Vlasios Vasileiou}
\affiliation{%
\institution{Experian DataLabs UK\&I and EMEA}
\city{London}
\country{United Kingdom}
}



\begin{abstract}
Understanding and removing bias from the decisions made by machine learning models is essential to avoid discrimination against \textit{unprivileged} groups. Despite recent progress in algorithmic fairness, there is still no clear answer as to which bias-mitigation approaches are most effective. Evaluation strategies are typically use-case specific, rely on data with unclear bias, and employ a fixed policy to convert model outputs to decision outcomes. To address these problems, we performed a systematic comparison of a number of popular fairness algorithms applicable to supervised classification. Our study is the most comprehensive of its kind. It utilizes three real and four synthetic datasets, and two different ways of converting model outputs to decisions. It considers fairness, predictive-performance, calibration quality, and speed of 28 different modelling pipelines, corresponding to both fairness-unaware and fairness-aware algorithms. We found that fairness-unaware algorithms typically fail to produce adequately fair models and that the simplest algorithms are not necessarily the fairest ones. We also found that fairness-aware algorithms can induce fairness without material drops in predictive power. Finally, we found that dataset idiosyncracies (e.g., degree of intrinsic unfairness, nature of correlations) do affect the performance of fairness-aware approaches. Our results allow the practitioner to narrow down the approach(es) they would like to adopt without having to know in advance their fairness requirements.
\end{abstract}

\maketitle

\section{Introduction}
The era of ``big data'' has seen industry seek competitive advantage through the monetisation of data assets and the use of modern machine learning methods in automated decisioning systems. However, there remain key questions on the impact these large-scale industrial algorithmic decisioning processes have on society ~\cite{WeaponsMathDestruction2016,AutomatedPredictionLaw2012}. One of these questions is: are these algorithms, which are often built from historic data with unknown sample and systematic biases, making fair decisions? This question has recently attracted significant attention, highlighted by media scandals on systematic unfairness presenting across many sectors including criminal recidivism~\cite{Recidivism2017}, recruitment ~\cite{HiringFairness2019,8937920} and credit-worthiness assessments. 

These cases have demonstrated that ``fairness through unawareness''~\cite{PhantomMenace2005,Dwork2012} is ineffective and misleading.  This realization spawned new research on developing new approaches to tackle unfairness in machine learning. Advances in the field now enable practitioners to specify the unprivileged groups of concern and employ fairness interventions at all points of the machine learning training process. Despite this progress, questions remain on the effectiveness of each intervention and how these techniques should be integrated into a single model-building framework. This is due to the number of potential sources of bias, the different and sometimes incompatible measures of fairness~\cite{FairnessImpossibility2018,Metrics2018, FairnessSurvey2019,Worldviews2019,Kusner2017,Chiappa2018}, the multitude of fairness interventions available, and the effects of decision policies employed to convert model outputs to decisions. The modeller must also address instances where fairness constraints and the machine learning optimisation criteria form competing goals and the imposition of fairness has a performance cost. 

In this paper, we address each of these problems and perform a systematic comparison of mainstream and convenient-to-use fairness interventions. In particular, we look for: 
\begin{itemize}
\item good quality, open-source code, 
\item documentation,
\item usability,
\item computational efficiency.
\end{itemize}
The comparison focusses on the case of binary classification with a single binary protected attribute. This protected attribute usually, but not exclusively, cannot be discriminated against by law.
\footnote{For example, see UK's 2010 Equality Act~\cite{ukequal2010}.} 

To facilitate the comparison, we introduce the metric \textit{fair efficiency} aimed to remove the effects of the decision threshold policy and to quantify the fairness-performance trade-off in a single value. Such a metric allows for a comparison of methods along a single axis, thereby solving one of the key issues in measuring the performance of fairness interventions. 

Finally, for comparison's sake, we propose a novel, simple approach called \textit{Fair Feature Selection} that selects features taking both their predictive power and fairness into account. 

The structure of this paper is as follows: in Section~\ref{sec:related_work} we provide additional background and further motivate our approach. In Section~\ref{sec:eval_fairness} we introduce our novel metric Fair Efficiency, section \ref{sec:methods} covers experimental methodology with results in Section \ref{sec:results} and discussion in~\ref{sec:discussion}. Our conclusions are then stated in Section~\ref{sec:conclusions}. Finally, \textit{Fair Feature Selection} is described in the Appendix. 

\section{Related work}
\label{sec:related_work}
Advances in algorithmic fairness and the development of interventions have correspondingly resulted in several fairness evaluation and comparative works. Interventions can broadly be categorised into: preprocessing of data~\cite{Reweighing2012, DIRemover2015, OptPreProc2017, Dwork2013}, post-processing of the model outputs~\cite{EqOddsPP2016, CalibEqOdds2017, RejectOptClf2012} and incorporating fairness constraints directly into model training ~\cite{IBM2018, FNN2018, Goel2018, MetaLearn2019, NaiveBayesFair2010, FairRed2018} (i.e., in-training). The incorporation of fairness constraints into model training often take the form of convex regularizations~\cite{FNN2018,Beutel2019,FairBoost2020}, adversarial learning~\cite{IBM2018, ShapSqueeze2020} or recasting the model-training process as one of constrained optimization~\cite{Zafar2017a,Zafar2017b,Zafar2017c, Olfat2018SpectralAF}. Other approaches utilize generative (or at least partly generative~\cite{FairInference2018}) models to identify and remove path specific effects to ensure counterfactual fairness~\cite{Kilbertus2017,Kusner2017,Chiappa2018,FairInference2018,CFGAN,ZhangDirectEffect}. We do not examine these latter generative approaches in detail. Instead, our study focusses on the more common use case of discriminative modelling in combination with statistical measures of group fairness. 

Ref.~\cite{FairnessSurvey2019} provides a broad survey of the bias sources and fairness in machine learning, creating a taxonomy for fairness definitions and analysing domain specific case-studies. A general methodology for exploring potential dataset biases was developed in~\cite{FairTest2019}. Ref.~\cite{Themis2017} designed test cases to automatically identify sources of discrimination in black-box decisioning procedures. Similarly,~\cite{FairMeasures2017} provides a framework to test underlying machine learning algorithms using a variety of datasets, fairness measures, and statistical tests. These works provide testing and evaluation metrics but are not designed for evaluative comparison of multiple fairness interventions at an equal footing, like this this work. 

Ref.~\cite{FriedlerComparison} evaluates the effects of data preprocessing, train-test splits and the formulation of fairness criteria in the algorithmic interventions. However, this work does not explicitly account for the role of the decision-threshold policy that is typically applied to the outputs of models. Furthermore, the comparison between interventions was point wise in nature and ignored how the metrics changed as the fairness hyper-parameters changed.

\section{Evaluating Fairness}
\label{sec:eval_fairness}
\subsubsection{Notation}
Consider a modelling dataset \linebreak $S = \{(x_1, z_1, y_1), \ldots,  (x_n, z_n, y_n)\} \subset \mathcal{X} \times \mathcal{Z} \times \mathcal{Y}$ where $(x_i, z_i, y_i)$ is the $i$th tuple respectively of: features, binary protected attribute (i.e., $\mathcal{Z} = \{0, 1\}$), and binary target label (i.e.,  $\mathcal{Y} = \{0, 1\}$). When $z=0$ the individual is in an unprivileged group, and when $z=1$ they are in the privileged group. An algorithm $\mathcal{A}: \mathcal{X} \rightarrow [0, 1]$ predicts a score $\hat{Y}=\mathcal{A}(x)$ which is compared to a threshold $\tau$ to assign binary labels $\bar{Y}$ to each dataset tuple, i.e. the predicted label is given by $\bar{Y} = \mathcal{I}(\hat{Y} > \tau)$, where $\mathcal{I}$ is the indicator function. Our convention is that $\bar{Y}=1$ represents the \textit{favourable} outcome. Examples of favourable outcomes include a loan application or a job application is approved, a credit limit is increased, one's bail application is approved, or one's parole application is approved. 

Finally, $\mathcal{A}$ may have a continuous or binary fairness parameter $\lambda$ which trades off fairness and classification performance. A value of $\lambda=0$ is utilized here to correspond to no fairness interventions, while a value of $\lambda=1$ corresponds to maximal strength of such interventions.

\subsubsection{Fairness Metrics}
Our aim is to measure two quantities: a predictive-performance metric quantifying how well the output scores predict the true labels, and also a fairness metric to determine how symmetric (i.e., unbiased or unprejudiced) the predicted labels are with respect to the protected attributes. The two classical fairness metrics we use are \emph{Equality of Opportunity} (EO) \cite{roemer2015equality} and \emph{Disparate Impact} (DI)
\footnote{Also referred to as ``Adverse Impact.'' For more information, see 
\href{https://en.wikipedia.org/wiki/Disparate\_impact}{https://en.wikipedia.org/wiki/Disparate\_impact}}. 

Equality of Opportunity is defined as:
\begin{displaymath}
\mathrm{EO} = 1 - |\mathrm{TPR}_{z=0} - \mathrm{TPR}_{z=1}|,
\end{displaymath}
where $\mathrm{TPR}_{z=0}$ and $\mathrm{TPR}_{z=1}$ are the true positive rates (proportion of correctly predicted positives) of the unprivileged and privileged classes, respectively. 

Disparate Impact is defined as: 
\begin{displaymath}
DI= min\left(\frac{P(\bar{Y} = 1| Z = 0)}{P(\bar{Y} = 1| Z = 1)}, \frac{P(\bar{Y} = 1| Z = 1)}{P(\bar{Y} = 1| Z = 0)}\right),
\end{displaymath}
where $P(\bar{Y} = 1| Z = 0)$ is the probability of a positive prediction for the unprivileged group, and $P(\bar{Y} = 1| Z = 1)$ is the corresponding probability for the privileged group. 

These are two of the most important fairness metrics, with the latter also having particular legislative significance in the United States. \cite{barocas2016big}. Typically, a DI value of less then $80\%$ is taken as an indicator of unwarranted discrimination and initiates further investigation (the ``four-fifths rule''). This rule originates from the Uniform Guidelines on Employee Selection Procedures, whereby a selection process will be deemed unfair if the success rate of the disadvantaged group is less than 80\% of the success rate of the advantaged group.\footnote{See \url{https://www.eeoc.gov/laws/guidance/questions-and-answers-clarify-and-provide-common-interpretation-uniform-guidelines.}} In many cases, DI is not optimized directly but through a related metric known as \textit{Statistical Parity Difference} (SPD), defined as the difference in probability of a positive prediction between the two groups, i.e. 
\begin{displaymath}
SPD = |P(\bar{Y}=1 | Z=1) - P(\bar{Y}=1 | Z=0)|. 
\end{displaymath}

\subsection{Fair efficiency} 
\label{sec:fair_efficiency}

One central question naturally arises: how can we benchmark $\mathcal{A}$ without \textit{a priori} knowledge of the appropriate $\lambda$ and classification thresholds $\tau$? One can either assume a specific combination of $\tau$ and $\lambda$. Alternatively, they can utilize a ``policy-agnostic'' approach, which can be facilitated using our novel metric \textit{fair efficiency}. 

Let us introduce a helper integral $K_m$, 
\begin{displaymath}
\label{eq:fe}
\begin{aligned}
K_m = \int_0^1\int_0^1 m(\lambda, \tau) d\tau d\lambda, 
\end{aligned}
\end{displaymath}
defined for any metric $m$ measuring fairness or predictive power. This integral considers all possible values of $m$ corresponding to the full range for all combinations of $\tau$ and $\lambda$. For the cases where $\lambda$ is discrete, the integral is replaced by an appropriate weighted sum over the values of $\lambda$. For the case of fairness-unaware algorithms, which do not have a $\lambda$, $K_m$ can be replaced with a single-point estimate. The metrics $m$ must be scaled so that they range from $[0,1]$ and that optimal performance corresponds to a value of $1$. 

Now, consider a predictive-performance metric $p$ (e.g., accuracy) and a fairness metric $f$ (e.g., DI or EO). How can we jointly evaluate $p$ and $f$ across all values of $\tau$ and $\lambda$? To accomplish that, we introduce the \textit{fair efficiency} $\Theta$ defined as the harmonic mean between $K_p$ and $K_f$, as:
\begin{equation}
\label{eq:fe_harmonic}
\begin{aligned}
\Theta_{p, f} = 2 \frac{K_p K_f}{K_p + K_f}.
\end{aligned}
\end{equation}
Fair efficiency penalises models that score highly for fairness but are not predictive, and vice versa. If the fairness and performance metrics are scaled so that a value of 0 (1) is maximally unfavourable (favourable), then $\Theta$ also follows the same pattern. Specifically, a $\Theta=0$ occurs when a model is either maximally unfair or maximally non-predictive, whereas an optimal $\Theta=1$ happens when the model is maximally predictive \textit{and} fair.

\section{Experimental Setup}
\label{sec:methods}
The aim of this evaluation is to compare a selection of practical fairness approaches across a diverse range of benchmark datasets. We will use standard metrics such as EO and DI, precision, accuracy, the area under the Receiver Operating Characteristic curve (AUC), as well as our policy-agnostic fair efficiency. 

We consider algorithms that are readily available for practitioners to use. This means they are open source, have good documentation, are able to run efficiently, and have adopted or are easily adapted to a standard fit/predict API. The latter criteria ensure that model selection and evaluation are easy to conduct in practice. We further require that the available implementations must have the ability to output a continuous score and that the fairness intervention must be controlled by a single continuous or binary parameter, $\lambda$. The former requirement excludes the MetaFairClassifier~\cite{MetaLearn2019, bellamy2018ai}, ART Classifier~\cite{bellamy2018ai} and the GerryFairClassifier~\cite{GerryFair2018} from the AI Fairness 360 (AIF360~\cite{bellamy2018ai}) library. Other algorithms such as Fair representations~\cite{Dwork2013, bellamy2018ai} violated the latter requirement or their implementations were not very usable, at the time of testing. Finally, we only evaluate fairness interventions applied at preprocessing and training stages, this puts post-processing interventions out-of-scope for this work. 

The first set of algorithms we consider are common fairness-unaware algorithms: logistic regression (Benchmark LR), Boosted Tree (Benchmark BT) naive Bayes classifier (Benchmark NB), Support Vector Machines (Benchmark SVM)~\cite{boser1992}, random forests~\cite{breiman2001random} (Benchmark RF), and boosted tree ensembles, such as XGBoost \cite{XGBoost2016} (Benchmark XGB) and LightGBM~\cite{ke2017lightgbm} (Benchmark LGB). These algorithms form a benchmark for achievable performance and fairness on our datasets without applying any fairness interventions. They're optimized across a range of hyperparameters in this comparison, e.g. logistic regression is considered in its unregularized form and in combination with l1, l2 and elastic-net regularization.

The fairness preprocessing techniques we consider are combined with a variety of mainstream classifiers. Specifically, we examine instance reweighing~\cite{Reweighing2012} and the disparate impact remover~\cite{DIRemover2015} (DIRemover). The Reweigher is taken directly form AIF360 while the DIRemover is an equivalent implementation in AIF360 but with a tunable $\lambda$. Reweighing is a two-state fairness algorithm, in which $\lambda \ge 0.5$ is equivalent to full reweighing and conversely $\lambda < 0.5$ indicates no reweighing. Although technically binary in nature, we still vary $\lambda$ across the full range and compute the integrated metrics as described in Section~\ref{sec:fair_efficiency}.  In addition, we also consider our novel, fair feature-selection method (see Appendix~\ref{sec:fair_feature_selecton}) to reflect how industry practitioners may attempt to tackle unfairness and bias through appropriate feature selection. The DIRemover debiases data by constructing new features, $\bar{x}$, from completely debiased ``fair'' features $\tilde{x}$ and the original inputs $x$. The debiased data is then given by $\bar{x} = (1-\lambda)x + \lambda \tilde{x}$.

The in-training fairness algorithms we consider include (a) SPD-regularized neural networks~\cite{FairBoost2020} (FR $\mbox{NN}_{0001}$) , (b) Lagrange reduced boosted trees and logistic regression~\cite{FairRed2018} (LagRed BT and LagRed LR), and (c) logistic regressions with fairness-constraint optimization~\cite{Zafar2017a}\footnote{https://github.com/mbilalzafar/fair-classification} (DI LR 1 and DI LR 2). 

For LagRed BT and LR, the fairness violation tolerance is mapped to $1-\lambda$. When $\lambda=1$, no unfairness, as measured by DP or EO is allowed. Conversely, $\lambda=0$ corresponds to an unconstrained lagrange reduction model. The logistic regressions DI LR 1 and DI LR 2 define unfairness as the covariance between $Z$ and the signed distance to the decision boundary. We consider two optimization forms: one where we optimize the logistic regression loss with the covariance constrained to $\pm0.05$ and another where we minimize the covariance subject to the loss being within $(1+\lambda)$ of the optimal loss. DI LR 1 does not have a natural $\lambda$ parameter and so $\Theta$ simply captures the trade-off between fairness and performance over all thresholds but not regularisation parameters. 

The neural network (FR $\mbox{NN}_{0001}$) is 3-layer fully connected networks with the number of units in each layer selected during hyper-parameter optimisation, the subscript denotes the layers where fairness-regularization losses are applied. 
  
The complete list of machine learning algorithms and fairness interventions considered can be found in Table \ref{tab:algos}.

\begin{table*}[ht]

\centering
\begin{tabular}{||c | c | c |}
\hline
name & type & package\\ 
\hline\hline
\makecell{Logistic regression (Benchmark LR)} & benchmark & sklearn  \\
\makecell{Naive Bayes Classifier (Gaussian) (Benchmark NB)} & benchmark & sklearn  \\
\makecell{Support Vector Classifier (RBF Kernel) (Benchmark SVM)} & benchmark & sklearn  \\
\makecell{Random Forest Classifier (Benchmark RF)} & benchmark & sklearn  \\
\makecell{BoostedTreeEnsemble (Benchmark BT)} & benchmark & sklearn \\
\makecell{XGBoost (Benchmark XGB)}  & benchmark & xgboost  \\
\makecell{LightGBM (Benchmark LGB)} & benchmark & lightgbm  \\
\makecell{Reweighing + random forest (Reweigh RF)}& pre-pro (on/off) & aif360 \\ 
\makecell{Reweighing + sklearn Boosted Trees (Reweigh BT)}& pre-pro (on/off) & aif360 \\ 
\makecell{Reweighing + LightGBM (Reweigh LGB)}& pre-pro (on/off)& aif360 \\ 
\makecell{Reweighing + logistic regression (Reweigh LR)}& pre-pro (on/off)& aif360 \\ 
\makecell{Reweighing + Gaussian naive Bayes (Reweigh NB)}& pre-pro (on/off)& aif360 \\ 
\makecell{Reweighing + support vector classifier (RBF kernel) (Reweigh SVM)}& pre-pro (on/off)& aif360 \\ 
\makecell{Reweighing + XGBoost (Reweigh XGB)}& pre-pro (on/off)& aif360 \\ 
\makecell{DIRemover + random forest (DIR RF)} & pre-pro & n/a \\
\makecell{DIRemover + boosted trees (DIR BT)} & pre-pro & n/a \\
\makecell{DIRemover + LightGBM (DIR LGB)} & pre-pro & n/a \\
\makecell{DIRemover + logistic regression (DIR LR)} & pre-pro & n/a \\
\makecell{DIRemover + Gaussian naive Bayes (DIR NB)} & pre-pro & n/a \\
\makecell{DIRemover + support vector classifier (RBF kernel) (DIR SVM)} & pre-pro & n/a \\
\makecell{DIRemover + XGBoost (DIR XGB)} & pre-pro  & n/a \\
\makecell{Fair feature selection ($\mathrm{FS}_8$ and $\mathrm{FS}_{12}$)} & pre-pro & n/a \\ 
\makecell{NeuralNetwork + SPD regularisation (FR $\mathrm{NN}_{0001}$ DI)} & in-train & n/a \\  
\makecell{Lagrange reduced boosted trees (LagRed BT)} & in-train & fairlearn  \\
\makecell{Lagrange reduced logistic regression (LagRed LR)}& in-train & fairlearn \\
\makecell{DI optimized logistic regression with boundary covariance constraints (DI LR 1)}& in-train (on/off) &  fair-classification \\
\makecell{DI optimized boundary covariance logistic regression with loss constraints (DI LR 2)} & in-train & fair-classification \\
\hline
\end{tabular}
\caption{Summary of benchmark and fairness algorithms included. ``in-train'' means that fairness is enforced as part of the training process, and ``pre-pro'' means fairness is achieved through preprocessing of the training dataset. Package versions: aif360 0.2.2~\cite{aif360-oct-2018}, scikit-learn 0.21.3~\cite{scikit-learn}, fairlearn 0.2.0~\cite{dudk2020fairlearn}, xgboost 0.9.0, lightgbm 2.3.0.}
\Description{}
\label{tab:algos}
\end{table*}

\subsection{Datasets}
\label{sec:data}

\begin{table}[th]
\begin{tabular}{|| c | c | c | c | c | c ||}
\hline
Name  & \makecell{Rows (k)} & \makecell{Features} &  \makecell{$<y>$} & $z$ & SPD \\ [0.5ex]
\hline\hline
Titanic  & 1.3 & 22 & 0.37 & gender & 0.51  \\ 
German  & 1 & 21 & 0.70 & gender & 0.07  \\  
Adult  & 48.8 & 14 & 0.24 & gender & 0.19   \\
Adult (race) & 48.8 & 14 & 0.24 & race & 0.10   \\
\makecell{S-D}& 10 & 12 & 0.53 & z & 0.14 \\ 
\makecell{S-P}& 10 & 16 & 0.51 & z & 0.10 \\  
\makecell{I-D}& 10 & 18 & 0.50 & z & 0.14 \\
\hline
\end{tabular}
\caption{Summary of datasets utilized in our comparison. To give an estimate of the ``intrinsic'' degree of unfairness of those datasets, we report the SPD of their target. As can be seen, the datasets are more-or-less well balanced label-wise.}
\Description{} 
\label{tab:data}
\end{table}

We use a collection of 3 real and 4 synthetic datasets, summarised in Table~\ref{tab:data}. The Titanic dataset~\footnote{https://www.openml.org/d/40945} contains attributes of the passengers onboard the Titanic ship that wrecked in 1912. The target shows whether an individual survived. The Adult dataset~\footnote{http://archive.ics.uci.edu/ml/datasets/Adult} contains socio-economic records on individuals to determine whether they earn over \$50K a year. The German dataset~\footnote{https://archive.ics.uci.edu/ml/datasets/statlog+(german+credit+data)} comprises features describing financial information of a set of individuals, the target being an indicator of a future credit default.

\begin{figure}
\begin{center}
\includegraphics[width=0.9\columnwidth] {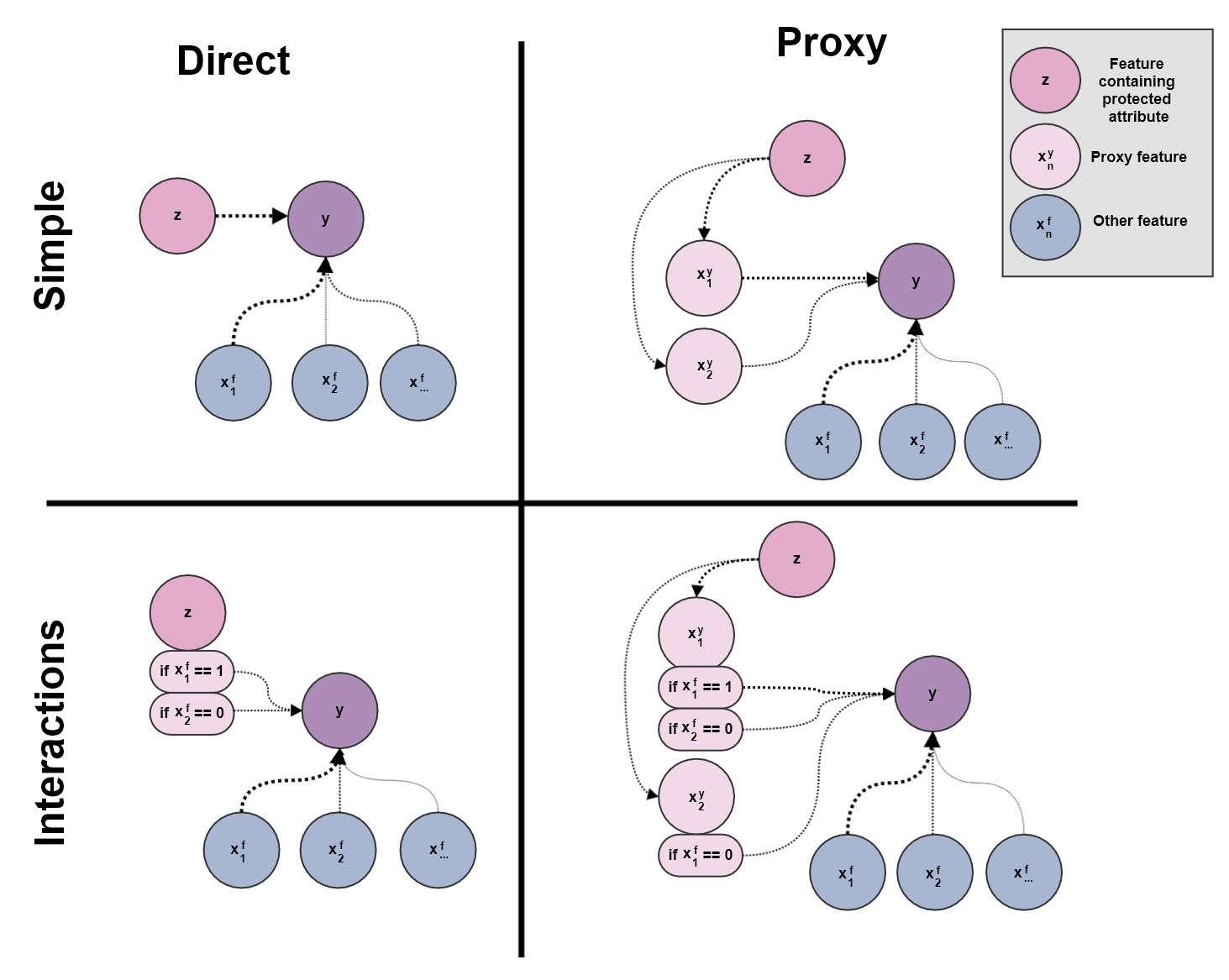}
\end{center}
\caption{Generation schemes for the synthetic datasets. Note that line weights are an illustrative indication that coefficient weights can vary, but do not necessarily represent the weights in the actual datasets.}
\Description{} 
\label{fig:methods_data_gen}
\end{figure}

Our synthetic datasets S-D, S-P, I-D, and I-P are generated according to Figure \ref{fig:methods_data_gen}. A linear scheme is used to first sample protected attribute $z$ from a Bernoulli distribution; followed by sampling $x$ in various ways to reflect different kinds of correlation between the training covariates $x$ and $z$. Next, a latent log-odds variable $s$ linearly combines the covariates $x$ and the protected attribute $z$, with intercept $s_0$. Finally, we sample $y \sim 1/(1 + \mathrm{e}^{-s})$. The above procedure gives rise to four different datasets: 
\begin{itemize}
 \item \textbf{Simple - Direct (S-D)} This dataset has a direct effect of $z$ on $y$, but no mediating effect through the $x$ variables. We sampled $x \sim \mathcal{N}(0, 1)$ and set $s = s_0 + \sum_i w_i x_i + w_z z$, where the sum is over the scalar variables $x_i$.
 \item \textbf{Simple - Proxy (S-P)} This dataset has no direct effect of $z$ on $y$, with the effect being instead mediated through some of the $x$ variables. We split $x$ in a set of ``fair'' variables $x^{(F)} \sim \mathcal{N}(0, 1)$ and a set of ``unfair'' variables $x^{(U)} \sim \mathcal{N}(z, 1)$, and set once again $s = s_0 + \sum_i w_i x_i$.
 \item \textbf{Interactions - Direct (I-D)} The direct effect of of $z$ on $y$ is turned on by a set of binary interaction variables. Splitting $x$ into a set of fair variables $x^{(F)} \sim \mathcal{N}(0, 1)$ and interaction variables $x^{(I)}$ sampled from a Bernoulli distribution, we set $s = s_0 + \sum_i w_i x^{(F)}_i + w_z z \prod_i x^{(I)}_i$ 
 \item \textbf{Interactions - Proxy (I-P)} This case is similar to the Simple - Proxy, but the effect of the unfair variables here is turned on by binary interaction variables. For simplicity, we used a single binary interaction variable $x^{(I)}$ sampled from a Bernoulli distribution. We therefore have $s = s_0 + \sum_i w^{(F)}_i x^{(F)}_i + x^{(I)} \sum_i w^{(U)}_i x^{(U)}_i$.
\end{itemize}

Although we do not characterise exactly how the bias is generated in the real datasets, a proxy for its magnitude is given by the target's SPD (see Table~\ref{tab:data}). For the Adult dataset, we evaluate the algorithms twice using two distinct protected attributes, race and gender. 

\subsection{Methodology}
\label{sec:EXP}

\begin{table}[t]
\begin{tabular}{||c | c | c | c||}
\hline
Dataset & train prop & test prop & no. reps \\ [0.5ex]
\hline\hline
Titanic and German & 0.5 & 0.5 & 5 \\ 
Adult & 0.2 & 0.8 & 3 \\
Synthetic & 0.4 & 0.6 & 4 \\ [1ex] 
\hline
\end{tabular}
\caption{Training/test splits and number of repetitions.}
\Description{}
\label{tab:fe_exps}
\end{table}

\subsubsection{Decision Policies}
\label{subsubsec:classsification_thresholds}
As shown in Section~\ref{sec:eval_fairness}, the fairness metrics DI and EO we will be using here are defined against a binary outcome $\bar{Y}$ produced after comparing the probability output of the algorithm with respect to classification threshold $\tau$. The choice of $\tau$ (``the policy'') is important, as it affects both fairness and predictive-performance metrics. In this work, we perform comparisons using three approaches:
\begin{enumerate}
 \item \textbf{``Argmax''}: We use a fixed threshold of $\tau_{Argmax}=0.5$ for all models and datasets. 
 \item \textbf{``PPR''}: We determine a threshold such that the positive predictive rate (PPR), (i.e., the likelihood of a favourable outcome) matches a pre-determined value of $20\%$ within a fixed tolerance.
 \item \textbf{Policy Free}: We compare algorithms treating all possible classification thresholds equally. This is accomplished using our fair efficiency metric. 
\end{enumerate}

For the first two policies we use accuracy and precision to measure predictive performance, whereas for the latter Policy-Free approach we use AUC. Given AUC is already integrated over $\tau$, the associated fair efficiency is given by:
\begin{equation}
K_{\rm{AUC}}= \int_0^1 (2*AUC(\lambda)-1) d\lambda.
\end{equation}

\subsubsection{Evaluation flow}
To evaluate the approaches, the individual datasets are split into train/test sets multiple times, see Table~\ref{tab:fe_exps}.
For each train/test repitition, we fit fairness-aware models as follows:

\begin{enumerate} 
\item \textbf{Optimization of ML-Algorithm-specific hyperparameters:} Using the training set, the optimal hyper-parameters for the model are determined using 3-fold cross-validation for $\lambda=0$ and utilizing AUC as the predictive-performance metric.

\item \textbf{Model Training:} 
    \begin{itemize}
    \item For all fairness aware algorithms, we sweep $\lambda$ from $0\rightarrow1$ and train models at each value.
    \item The models are trained using the optimal hyper-parameters identified in Step 1. where possible.
    \end{itemize}
\item \textbf{Evaluation of fairness:}
    \begin{itemize}
    \item For all trained models, the thresholds $\tau$ are swept from $0\rightarrow1$.
    \item The predictive-performance and fairness metrics are computed for each $\tau$ and $\lambda$.
    \item The predictive-performance and fairness integrals $K_{\rm{AUC}}$ and $K_{\rm{DI}}$/$K_{\rm{EO}}$ are first calculated on the test set for each value of $\lambda$, and then they are combined using a harmonic mean to produce the fair efficiencies $\Theta_{\rm{AUC,DI}}$ and $\Theta_{\rm{AUC,EO}}$. These metrics are independent of the threshold policy and utilize all $\lambda$ values.
    \end{itemize}
\end{enumerate}

We use $k=8$ or $12$ features for fair feature selection and use these to fit a random forest model ($\mathrm{FS}_8$ and $\mathrm{FS}_{12}$, respectively). A random forest is used as the estimator for its general utility and ease of fitting, and as it is also used in the other preprocessing intervention pipelines. For the benchmark models we use point estimates of efficiency at $\lambda=0$.

\subsubsection{Matching the Decision Policies and Experimental Comparisons}
\label{subsubsec:policiesandcomp}
The evaluation flow of the previous Section produces many metrics for each $(\tau, \lambda)$ combination. These metrics need to be extracted according to match the Argmax and PPR policies. The determination of $\tau$ for each model happens as follows:

\begin{itemize}
 \item For Argmax, we use $\tau_{argmax}=0.5$.
 \item For PPR, we scan across all $\tau$ and $\lambda$ combinations to find the smallest $\tau$ closest to the $20\%$ target acceptance rate. This scan is across all training dataset repititions. If the average acceptance rate across all repititions is not within $\pm3\%$ of $20\%$, then the model has not met the PPR criteria and is dropped. We do not report any further results for these dropped models. 
 \item For the Policy-Free approach, all values of $\tau$ are considered.
\end{itemize}

Given the appropriate $\tau$ and $\lambda$ combinations, for PPR and Argmax, we perform several comparisons.
The fairness preprocessing interventions are initially compared separately in an on/off manner. This approach allows us to examine the maximum expected fairness gains one can expect from these preprocessing techniques. As the DIRemover has a tunable fairness parameter, we compare this algorithm at maximum $\lambda$ in conjunction with the benchmark models to the performance of the benchmark models in isolation. The former case is referred to as the fairness ``on'' state and the latter fairness ``off''. Similarly we compare the instance reweigher plus a benchmark classifier at maximum and minimum $\lambda$.

For all fairness-aware interventions, and for the Argmax and PPR policies, we select a value of $\lambda$ that is consistent with a ``fairness budget''. Specifically, for each approach and dataset we identify the $\lambda$ value corresponding to both maximum accuracy and a DI~$>0.8$ on at least one training dataset repetition. Given these $\lambda$, we observe the mean DI/EO and performance metrics across the train/test repetitions. For the Policy-Free approach, which is facilitated by our fair efficiency metric, there is no particular choice of $\lambda$, as all $\lambda$ are being considered equally. 

Finally, unless stated otherwise, the reported metrics are averaged across the train/test repititions of each dataset. 

\section{Results}
\label{sec:results}

\subsection{Argmax and PPR Policies}

\subsubsection{Benchmark models}
\begin{figure*}[t]
\centering
  \hspace{0.05\linewidth}
  \includegraphics[width=1.05\textwidth]{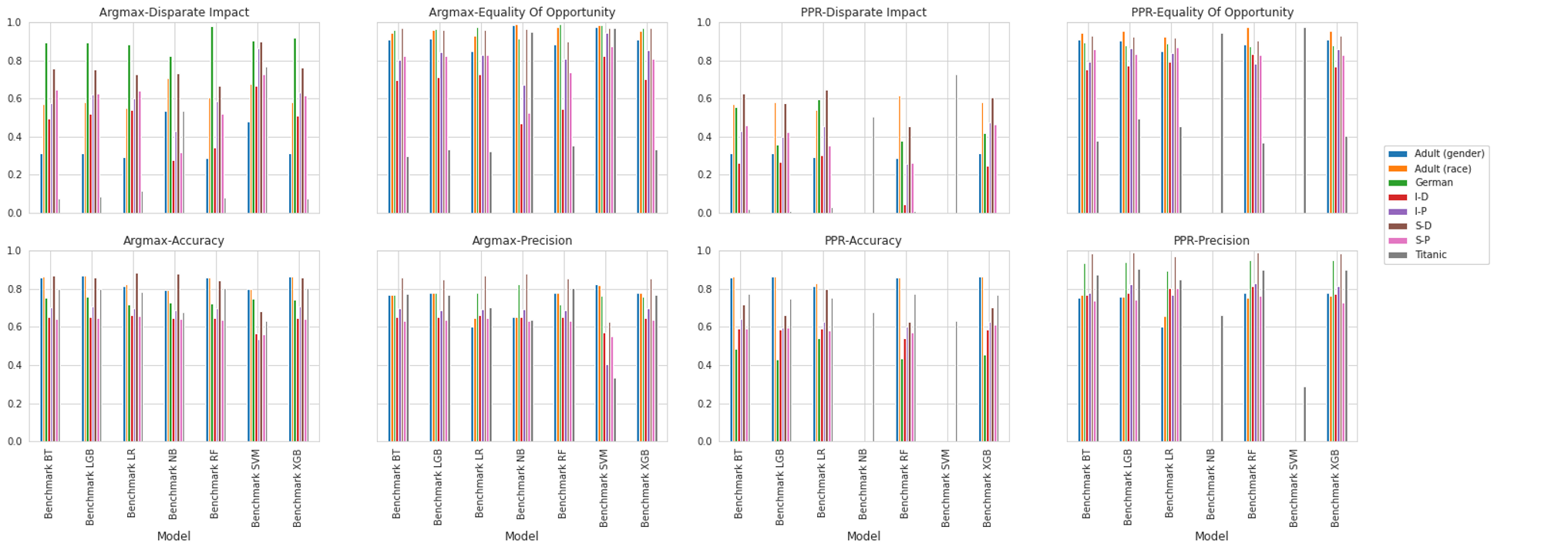}
  \caption{\label{fig:benchmarkperf}
  \Description{}
Fairness and performance metrics of benchmark models across all datasets under Argmax and PPR threshold policies.}
\end{figure*}

The benchmark models' precision, accuracy, DI and EO are shown in  Figure~\ref{fig:benchmarkperf}. As can be seen and for the most cases, the predictive powers of the benchmark models were more or less consistent (within $\pm 0.15$) given a particular dataset and policy. Fairness-wise, while EO was mostly consistent across models and datasets, the variations in DI across datasets were strong. Importantly enough, in the vast majority of cases, benchmark models failed to surpass the $80\%$ DI threshold, corresponding to the ``four-fifths rule''. 

Observing how the different algorithms performed, we notice that the simple Logistic Regression was not overall the fairest algorithm. On the contrary, and as reported in Table~\ref{tab:benchmark1} (for the case of Argmax policy), SVM was the overall fairest benchmark model. 

\begin{table}[t]
\begin{tabular}{||c | c | c | c||}
\hline
Dataset & Best Model-EO & Best Model-DI \\ [0.5ex]
\hline\hline
Titanic & Benchmark SVM & Benchmark SVM \\
German & Benchmark LR & Benchmark RF \\
Adult (gender) & Benchmark NB & Benchmark NB \\
Adult (race) & Benchmark RF & Benchmark NB\\
\makecell{S-D}&  Benchmark SVM & Benchmark SVM \\ 
\makecell{S-P}& Benchmark SVM & Benchmark SVM \\  
\makecell{I-D}& Benchmark SVM & Benchmark SVM \\
\makecell{I-P}& Benchmark SVM & Benchmark SVM \\[1ex]
\hline
\end{tabular}
\caption{Fairest benchmark models on each dataset using Argmax policy.}
\Description{} 
\label{tab:benchmark1}
\end{table}

\subsubsection{Preprocessing approaches}
\begin{figure*}
\centering
  \hspace{0.05\linewidth}
  \includegraphics[width=1.05\textwidth]{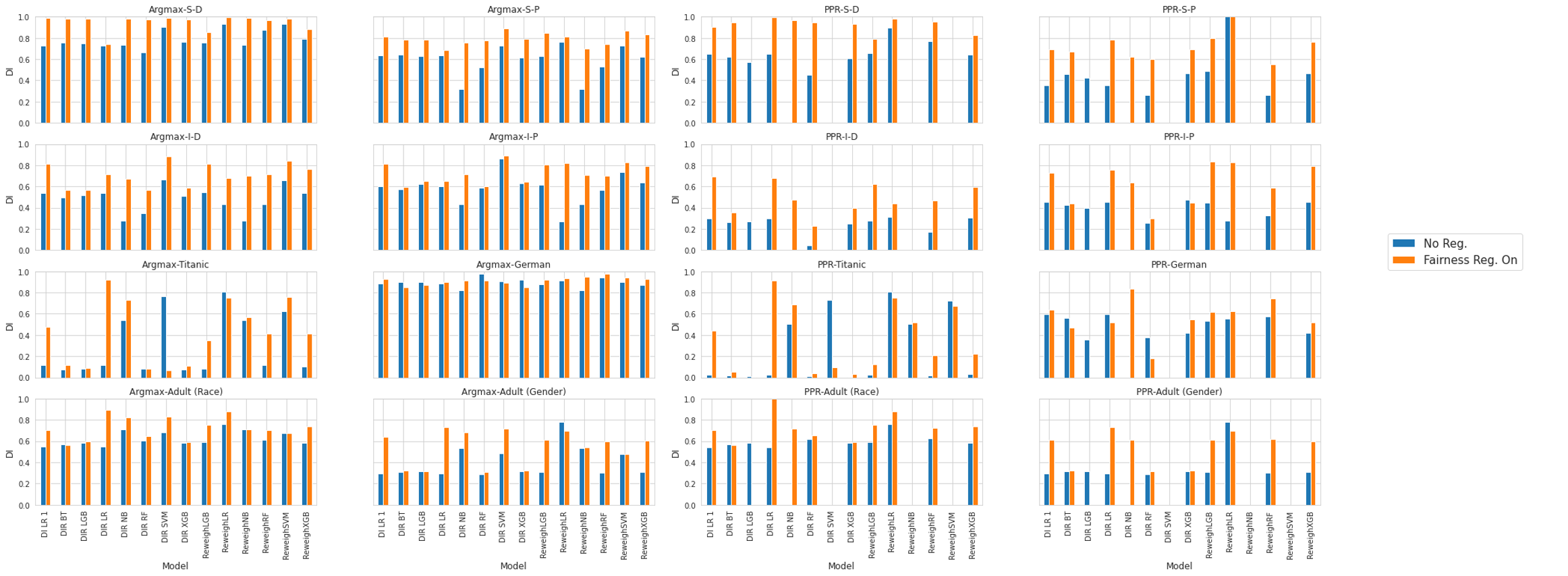}
 \caption{\label{fig:onoff_perf} DI for preprocessor interventions, in ``on/off'' comparison, across all datasets for Argmax and PPR on the test datasets.}
 \Description{} 
\end{figure*}

We begin with a binary comparison, see Section~\ref{subsubsec:policiesandcomp}, of the fairness preprocessing techniques, namely, DIRemover and Reweighing. For most combinations of policy, approach, and underlying machine learning algorithm we found a significant uplift in DI when the fairness intervention is applied, see Figure~\ref{fig:onoff_perf}. Comparing the uplifts across datasets or underlying methods or policies, we observe a high variance. In general, the Argmax policy allowed for more consistent uplifts across datasets and underlying ML algorithms. On the other hand, the uplifts in the PPR policy exhibit a more erratic  behaviour. We also observed that the two fairness approaches combined in similarly fruitful ways with our range of underlying ML algorithms. Finally, the two techniques, overall, produced similar levels of uplifts. 

We once again observed numerous instances where the fairness interventions do not result in DI~$>0.8$ (as recommended by the four-fifths rule). This is particularly problematic for the PPR policy, where the majority of fairness-optimized models are still below this DI-threshold. 

Contrastingly to DI, but similar to the benchmark models, we observed high values of EO ( $>0.8$) for the non-fairness-induced models, particularly in combination with the Argmax policy, and generally the fairness regularizations improved this metric further (not shown for brevity). This supports our initial observation that EO, depending on both the target and prediction, is easier to optimize than DI and effective learning of the target assists in making the model fairer.

Finally, we should mention that the uplifts in DI from both approaches came with tolerable accuracy and precision penalties (median drop in absolute value < 0.02).

\subsubsection{Overall performance}
\begin{figure*}
\centering
   \includegraphics[width=1.05\linewidth]{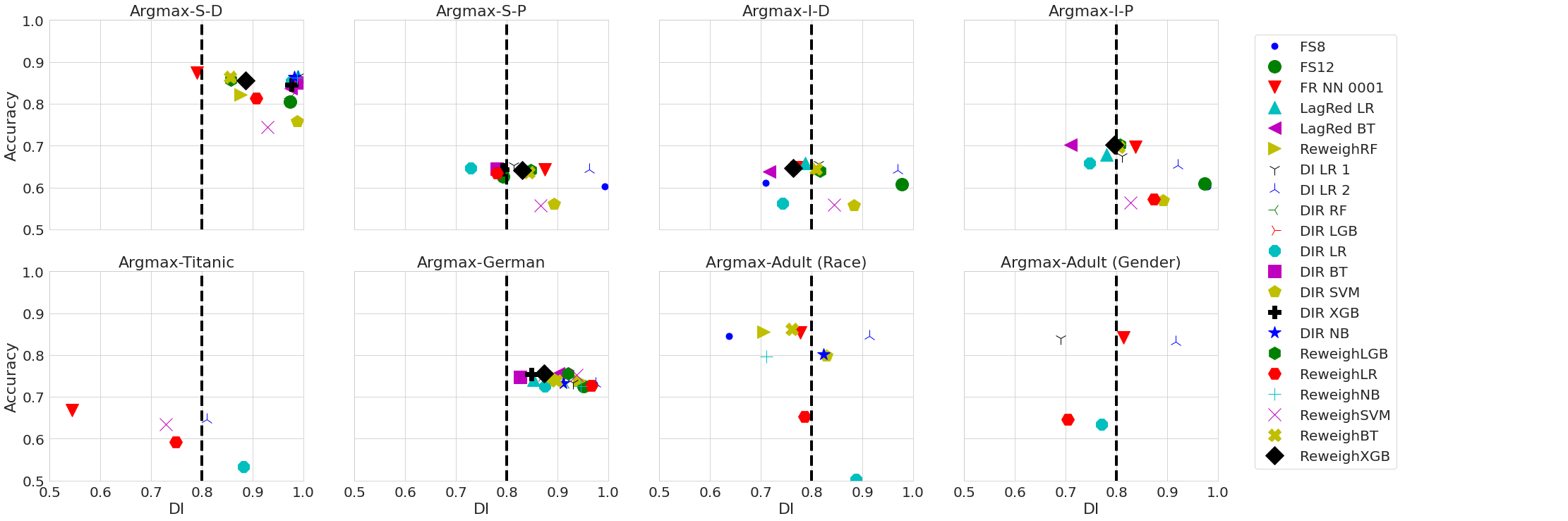}
 \caption{\label{fig:scatterperf} Accuracy and DI for all fairness approaches under Argmax}
 \Description{} 
\end{figure*}

We continue by considering both preprocessing and in-training fairness techniques, utilizing the accuracy and DI measures of performance. We note that in several of the cases, we could not find an appropriate decision threshold to satisfy both the PPR and/or the Argmax policies as well as the requirement of DI$>0.8$ -- with the situation being worse for the PPR policy.

Figure~\ref{fig:scatterperf} shows the results for the case of Argmax policy.  Observing within the datasets and for all datasets, except maybe the simplest S-D, we do not encounter a strong anti-correlation between fairness and accuracy -- something that may be be naively expected (in an industrial setting). Instead, we notice that one may obtain strong gains in fairness (DI gains up to $\sim0.3$) without sacrificing accuracy if multiple combinations of fairness interventions and underlying modelling approaches are trialled. Finally, we observed that in all datasets, there is at least one combination that can produce DI~$>0.8$ results.

\begin{table}[ht]
\resizebox{\columnwidth}{!}{
\begin{tabular}{||c | c | c | c | c | c||}
\hline
Dataset & Model & Accuracy & Precision & DI & EO \\ [0.5ex]
\hline\hline
German&Reweigh XGB&$0.76$&$0.78$&$0.87$&$0.95$\\
Titanic&DI LR 2& $0.65$&$0.53$&$0.81$&$0.92$\\
Adult (gender)&FR $\mathrm{NN}_{0001}$&$0.84$&$0.73$&$0.81$&$0.74$\\
Adult (race)&DI LR 2&$0.84$&$0.74$&$0.91$&$0.93$\\
S-D&LagRed LR&$0.87$&$0.85$&$0.99$&$0.89$\\
S-P&DI LR 1&$0.65$&$0.65$&$0.81$&$0.94$\\
I-D&DI LR 1&$0.66$&$0.66$&$0.81$&$0.93$\\
I-P&Reweigh LGB &$0.70$&$0.69$&$0.81$&$0.98$\\[1ex]
\hline
\end{tabular}}
\caption{Most accurate fairness approaches for DI~$>0.8$ under Argmax. Metrics associated with the maximum-accuracy test-set repetition are reported.}
\Description{} 
\label{tab:optim1}
\end{table}

\begin{table}[ht]
\resizebox{\columnwidth}{!}{
\begin{tabular}{||c | c | c | c | c | c||}
\hline
Dataset & Model & Accuracy & Precision & DI & EO \\ [0.5ex]
\hline\hline
German&FR $\mathrm{NN}_{0001}$&$0.57$&$0.89$&$0.80$&$0.94$\\
Titanic&Reweigh LR&$0.53$&$0.40$&$0.81$&$0.98$\\
Adult (gender)&DI LR 2&$0.82$&$0.63$&$0.92$&$0.68$\\
Adult (race)&DI LR 2&$0.82$&$0.66$&$0.95$&$0.92$\\
S-D&DI LR 2&$0.76$&$0.97$&$0.97$&$0.85$\\
S-P&DI LR 2&$0.61$&$0.73$&$0.91$&$0.93$\\
I-D&DI LR 2&$0.59$&$0.76$&$0.91$&$0.95$\\
I-P&FR $\mathrm{NN}_{0001}$&$0.60$&$0.77$&$0.89$&$0.91$\\[1ex]
\hline
\end{tabular}}
\caption{Most accurate fairness approaches for DI~$>0.8$ under \textbf{PPR}. Metrics associated with the maximum-accuracy test-set repetition are reported.}
\Description{} 
\label{tab:optim2}
\end{table}

To conclude this analysis, we identified the most accurate models (including benchmark models) on each dataset, where the DI was $>0.8$ on both the training and test set. These results are shown in Tables~\ref{tab:optim1} and ~\ref{tab:optim2}, for the Argmax and PPR policies, respectively.
In both cases, the optimal models were typically a member of the constrained optimization family, i.e., DI LR 1/2 or FR $\rm{NN}_{0001}$. As expected, no benchmark models ended up being the most accurate given a DI$>0.8$. As mentioned above, benchmark models typically struggled to reach this level of DI. 
A natural question that would arise is how much accuracy would one need to sacrifice to reach a ``compliant'' ($>0.8$) level of DI? We compare the benchmark models that give the largest accuracy on a test dataset repetition to the most accurate ``compliant'' model. For both PPR and Argmax, we observe small but varied drops in accuracies on the order of $\sim0.02-0.04$. The Titanic dataset is an exception showing drops of $\sim0.13$ and $\sim0.16$ for PPR and Argmax respectively.

We also consider the calibration of the models as measured by Brier score on the test data repetitions.  In all cases, the models identified as being the most calibrated were fairness-unaware (i.e., they were benchmark models). The optimal Brier scores were always $<0.22$ for these models. Moreover, on the Adult and Titanic datasets the difference between the most calibrated and least calibrated models (i.e. fairness-aware) was $>0.41$. This highlights that fairness and calibration requirements can often be in conflict, resulting in potential barriers to industry adoption where calibrated outputs are a necessity.

\subsubsection{Computational Cost}
\begin{figure}
\centering
     \includegraphics[width=\columnwidth] {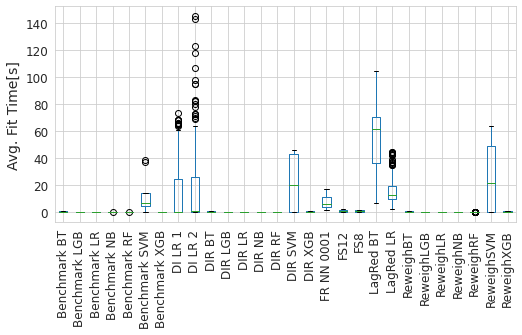}
     \caption{Box-plots of fit times across all datasets for benchmark and models with fairness interventions.} 
     \Description{} 
     \label{fig:timings}
\end{figure}
Finally, we consider the computational cost in training these models, tested on an Intel Xeon E5-2680 at 2.80GHz with 128 GB RAM. Taking the mean training times for each model and dataset across all $\lambda$ and repetitions considered, we observe a clear pattern in Figure~\ref{fig:timings}. The preprocessing interventions and benchmark models are the fastest to fit followed by the fairness-loss regularized models. The most expensive models to fit are those that employ meta-learning approaches using complex base learner, i.e., LagRed BT which had a mean fit time $\gtrsim 50$s, and fairness constrained learners. In particular DI LR 2 can often get ``stuck'' when trying to match the hard optimization constraints resulting in inefficient outlier runs of $>100$s. This highlights that although these models are consistent in their performance and fairness utility, they can come with a potentially prohibitive high computational cost.

\subsection{Fair efficiency}
\begin{figure}
\begin{center}
     \includegraphics[height=1.20\columnwidth] {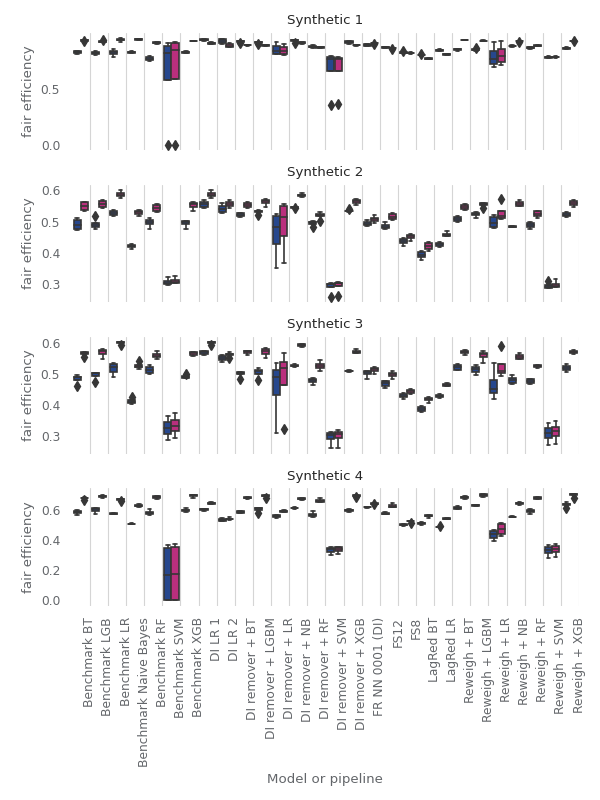}
     \includegraphics[height=1.20\columnwidth] {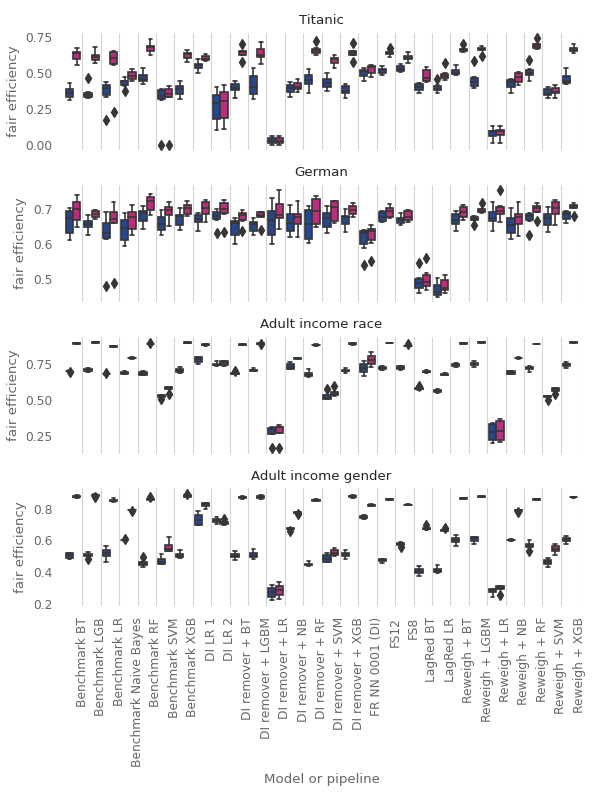}
\end{center}
     \caption{$\Theta_{\rm{DI}}$ and $\Theta_{\rm{EO}}$ for all our datasets and techniques.} 
     \Description{} 
     \label{fig:results_fe}
\end{figure}

We continue by utilising a policy-free approach facilitated by our fair efficiency $\Theta$. In the following, $\Theta$ is calculated twice for each model alternatively using DI and EO as the fairness component, which are presented as pairs in the box plots (DI on the left of each pair). Error bars represent variance across the folds of each dataset. $\Theta_{\rm{DI}}$ and $\Theta_{\rm{EO}}$ for all our datasets and techniques are shown in Figure~\ref{fig:results_fe}.

Comparing results across datasets, fairness metrics, and techniques, there are some interesting patterns we can observe. 

Evaluating $\Theta$ across techniques we observe overall consistency with only few underperforming outliers, namely SVM-based approaches (Benchmark SVM, DI Remover SVM, and Reweigh + SVM), Logistic Regression based approaches (Benchmark LR, DI Remover LR, and Reweigh LR), the two Lagrange-reduced techniques LagReg BT/LR. Finally, it is positive to see the benchmark models being competitive to all the other fairness-aware approaches. 

Focussing on LagRed BT and LagRed LR, which struggled across the datasets, we believe that this is likely due to the observed behaviour as $\lambda$ changes. Specifically, for these algorithms, $\lambda$ is an unfairness tolerance parameter that fails to control model fairness in a monotonic fashion (as ideally, expected). 

We observed fairness increasing over a small range of this parameter. This does not necessarily indicate that the algorithm is ineffective, but rather it may be harder to align the models' performance and fairness requirements. 
Comparing fair efficiencies for DI to those of EO, we observe that while the gap was relatively small in most synthetic datasets, the situation is different in three out of four real datasets. In general, $\Theta_{\rm{EO}}$ was higher than $\Theta_{\rm{DI}}$, implying that EO is easier to maximize than DI -- a finding in agreement with previous results.

\section{Discussion}
\label{sec:discussion}

Having a full view of the results across all decision-threshold policies, we proceed with describing the emerging picture and main points.

First of all, it was easier to obtain high values of EO rather than DI. While there is some guidance for the case of DI on which values correspond to an adequately fair model (four-fifths rule), there is no corresponding rule for EO. Is an $80\%$ EO fair? If so, then one may achieve fairness more easily by focussing on EO instead of DI. 
 
The vast majority of deployed models in the industry are using fairness-unaware algorithms. Our benchmark evaluations revealed non-compliance with the four-fifths rule on some/most cases, although the overall picture is not terrible. On the positive side, the fairness-vs-predictive-power aspects of benchmark models (as measured by $\Theta$) were comparable to those of fairness-aware models (see, Figure~\ref{fig:results_fe}). Moreover, for EO, benchmark models performed adequately well producing high values of EO~$\simeq 0.9$ or above for both Argmax and PPR policies (see Figure~\ref{fig:benchmarkperf}). On the other hand, the situation with regard to DI was less positive, with benchmark models failing to reach a compliant ($80\%$) level of DI in almost all cases. This indicates that that fairness-unaware algorithms may be reasonably subject to fairness concerns and investigations.
 
 Among the benchmark algorithms, and for our tested datasets and Argmax policy, SVM had overall the best fairness profile (see Table~\ref{tab:benchmark1}). However, SVM's predictive power was not as high as that of other algorithms (e.g., see Figures~\ref{fig:benchmarkperf} and ~\ref{fig:results_fe}).
 
On the topic of benchmark models, we would also want to evaluate the widely spread (at least in industrial circles) belief that ``simpler models are fairer'', which may stem partly from the assumption that more complex models are harder to explain.  On this basis, the simple logistic regression benchmark could be assumed to be the most fair, with the less-transparent boosted tree approaches (e.g., Benchmark XGB) be the least fair. Our results, however, show this is not the case, as we did not observe a clear anti-correlation between simplicity and fairness. This result is reassuring and means complex models should not necessarily be precluded by fairness concerns.
 
How could someone improve the fairness of their model while using a fairness-unaware algorithm? One approach would be to try different decision-threshold policies. One can always modify their threshold $\tau$ until a satisfactory fairness level is achieved. However, this approach may have other material, business-related side effects associated with it. For example, in a loan application, this approach might result into approval rates incompatible with business constraints.
 
On the other hand, our results show that fairness-aware algorithms are effective in finding effective trade-offs between fairness and performance. As we have seen in Figures~\ref{fig:onoff_perf} and \ref{fig:scatterperf}, one can gain appreciable amounts of fairness with minimal drops in predictive power. Moreover, it is positive to see that compliant levels of DI are routinely achievable using such algorithms (e.g., see Tables~\ref{tab:optim1} and \ref{tab:optim2}, and Figure~\ref{fig:onoff_perf}). 
 
However, which family of fairness-aware algorithms fared better? Were preprocessing or in-training approaches better? Were on-off (binary) approaches adequate, or should users focus on approaches having a continuous configurable parameter? 
 
Considering the results in Tables~\ref{tab:optim1} and \ref{tab:optim1}, we saw that the two in-training approaches (DI LR 1 and DI LR 2) from the fair-classification package provided the most accurate-yet-fair models for about half of the combinations tested across Argmax and PPR. Looking at the scatter plot in Figure~\ref{fig:scatterperf}, we also notice these two techniques being competitive in terms of absolute levels of DI (for the Argmax policy). Moreover, their fair efficiency profiles were overall competitive, too (see Figure~\ref{fig:results_fe}).
 
We notice that both DI LR 1 and DI LR 2 are ``in-train'' type of fairness algorithms. Is this family of algorithms, better than preprocessing ones? Our results did not reveal any obvious pattern performance-wise, as the other major player in in-train algorithm family, the pair of LagReg BT and LagReg LR, performed poorly with respect to fair efficiency (see Figure~\ref{fig:results_fe}). However, performance aside, it should be mentioned many implementations of preprocessing techniques (e.g. the reweigher) are binary (on/off), and, as such, they do not offer as much flexibility as some of the fully-configurable in-train approaches (i.e., FR $\rm{NN}_{0001}$ DI, LagRed LR, and DI LR 2). 
 
Unsurprisingly we found that the benchmark models were better calibrated than those with a fairness intervention. The fairness-aware models were poorly calibrated in comparison to the benchmark models and any additional requirements on calibration could negatively impact the fairness properties of the models and make them inviable~\cite{FairnessImpossibility2018}. Any additional requirements on calibration, may require post-processing interventions to account for this constraint.

The two Lagrange reduction approaches (LagRed BT and LagRed LR) were found to perform poorly compared to other interventions considered. With the exception of the Titanic dataset, these models could not meet the PPR requirements. Furthermore, their fair efficiency was typically lower than those of the other models considered. We attribute this to form of the fairness constraint and evaluation methods. For these models, fairness was peaked over a small range of $\lambda$ and low outside this range, thus lowering its associated $\Theta$. 

Finally, the similar performance across the synthetic datasets indicates that the tested interventions function independently of the structure of the bias in the data. This is unsurprising, as the interventions focus on metrics that are agnostic of the underlying structure of the data-generation mechanism. Other metrics that do account for these mechanisms, such as the causal losses defined in ~\cite{FairBoost2020}, may offer a better fairness-performance trade-off than those considered in this study.

\section{Conclusions}
\label{sec:conclusions}

We evaluated 28 different models, comprising 7 fairness-unaware (benchmark) ML algorithms and 20 fairness-aware models driven by 8 fairness approaches. We utilized 3 decision threshold policies, 7 datasets, 2 fairness metrics, and 3 predictive-performance metrics. 

This is the most comprehensive comparison in the literature to-date and uses a new approach to generalising the evaluation metrics, using our novel metric fair efficiency. Fair efficiency is usable within existing experimental frameworks and provides an alternative view on the trade-off between performance and fairness to context specific metrics. 

We also introduced a new feature selection technique that was comparable to the other preprocessing techniques considered and in some instances was competitive with in-training techniques. 

The comparisons presented here are by no means exhaustive, and are limited by our strict criteria for inclusion. As of yet, there is no standard ``model-fairness API''. Disparate APIs and functions, add significant overhead to comparing models, and we hope that more algorithms will move into the scope set by this paper as consensus emerges.

Future work can expand the scope of this comparison to include causal techniques, additional fairness metrics, post-processing techniques, model interpretability and common business constraints such as calibration.

\appendix
\section{Fair feature selection}
\label{sec:fair_feature_selecton}
\textit{Fair feature selection} is a straightforward fairness-induction approach used here as a baseline to be compared to more evolved algorithms. It reduces prediction bias by selecting features based on how well they predict $y$, penalised by how well they predict $z$. Features are ranked in two stages. First the features are scored against $y$ using a weighted combination of the ranks calculated using three selection methods; mutual information (MI, weight $W$=0.15), F-test score (FS, $W$=0.15), and the gain importance from a random forest model (GI, $W$=1). Next the target is set to the $Z$ and each feature scored again using the same weighted combination of MI, FS and GI. The weighting between $y$ predictive power and the inverse-$z$ predictive power is controlled by a $\lambda$ parameter. At $\lambda=0$ feature ranking is based on $y$ predictive power only and at $\lambda=1$ ranking is based on a feature's inverse-power to predict $z$ only. The top $k$ ranked features based on this rank are returned. See Algorithm \ref{alg:fair_feature_selecton} for the pseudo-code.

\begin{algorithm}
\caption{Fair feature selection}
\label{alg:fair_feature_selecton}
\begin{algorithmic}[1]
\REQUIRE Training data $\{(\mathbf{x}_1, z_1, y_1), \dots, (\mathbf{x}_n, z_n, y_n)\}$, output features $k$, ranking weights $\mathbf{w}$, mixing parameter $\lambda \in [0, 1)$ 

\STATE Initialize ranking vector $\mathbf{S}$.
\FOR {i = 1,\dots, len($\mathbf{x}$)}
\STATE Compute ranking score of $\{\mathbf{x}[i]\}$ using $y$:\\ $S_y=\mathbf{w}^{T}($MI, F-test score, GI$)^{T}$.
\STATE Compute ranking score of $\{\mathbf{x}[i]\}$ using $z$:\\ $S_z=\mathbf{w}^{T}($MI, F-test score, GI$)^{T}$.
\STATE Store the result: $\mathbf{S}[i] =  \lambda S_y + (1 - \lambda)S_z$.
\ENDFOR
\RETURN Top $k$ features from ranking
\end{algorithmic}
\end{algorithm}

\bibliography{main}
\bibliographystyle{ACM-Reference-Format}
\end{document}